\pdfoutput=1

\documentclass[11pt]{article}

\usepackage{acl}

\usepackage{times}
\usepackage{latexsym}
\usepackage{xcolor}

\usepackage[T1]{fontenc}

\usepackage[utf8]{inputenc}
\usepackage{mathalpha}
\usepackage{amsmath,amssymb}
\usepackage{booktabs}
\usepackage{multirow}
\usepackage{adjustbox}
\usepackage{CJKutf8}
\usepackage{tipa}

\usepackage{microtype}

\title{\textbf{\textsc{Readin}}: A Chinese Multi-Task Benchmark with \\
Realistic and Diverse Input Noises}

\author{\hspace{0.3cm} Chenglei Si$^{1,2*}$ \hspace{0.7cm} Zhengyan Zhang$^{1*}$ \hspace{0.3cm} Yingfa Chen$^{1*}$ \\
 \textbf{Xiaozhi Wang}$^{1}$ \hspace{1.2cm} \textbf{Zhiyuan Liu}$^{1,3\dagger}$ \hspace{0.8cm}  \textbf{Maosong Sun}$^{1,3\dagger}$ \\
  $^1$ NLP Group, DCST, IAI, BNRIST, Tsinghua University, Beijing \\
  $^2$ University of Maryland, College Park, MD, USA \\
  $^3$ International Innovation Center of Tsinghua University, Shanghai\\
         \texttt{clsi@terpmail.umd.edu, zy-z19@mails.tsinghua.edu.cn}\\
}

\begin{document}
\maketitle

\begin{CJK*}{UTF8}{gbsn}
\begin{abstract}

For many real-world applications, the user-generated inputs usually contain various noises due to speech recognition errors caused by linguistic variations\footnote{Note that linguistic variations themselves are not noises or errors, but they can lead to noises in the data processing for example due to failure of speech recognition.} or typographical errors (typos). 
Thus, it is crucial to test model performance on data with realistic input noises to ensure robustness and fairness. 
However, little study has been done to construct such benchmarks for Chinese, where various language-specific input noises happen in the real world. 
In order to fill this important gap, we construct \textbf{\textsc{Readin}}: a Chinese multi-task benchmark with \textbf{\textsc{RE}}alistic \textbf{\textsc{A}}nd \textbf{\textsc{D}}iverse \textbf{\textsc{I}}nput \textbf{\textsc{N}}oises. 
\textsc{Readin} contains four diverse tasks and requests annotators to re-enter the original test data with two commonly used Chinese input methods: Pinyin input and speech input.
We designed our annotation pipeline to maximize diversity, for example by instructing the annotators to use diverse input method editors (IMEs) for keyboard noises and recruiting speakers from diverse dialectical groups for speech noises. 
We experiment with a series of strong pretrained language models as well as robust training methods, we find that these models often suffer significant performance drops on \textsc{Readin} even with robustness methods like data augmentation. 
As the first large-scale attempt in creating a benchmark with noises geared towards user-generated inputs, we believe that \textsc{Readin} serves as an important complement to existing Chinese NLP benchmarks.
The source code and dataset can be obtained from \url{https://github.com/thunlp/READIN}.


\end{abstract}

\section{Introduction}

{\let\thefootnote\relax\footnotetext{$^*$ Equal contribution}}
{\let\thefootnote\relax\footnotetext{$^\dagger$ Corresponding authors}}


User-generated inputs in real-world applications often contain noises where wrong characters or words are used instead of the intended ones~\cite{2021ProceedingsOT}. This is especially true when users type fast or are using speech input in noisy environments or with less common accents that cause errors in post-processing systems. 
However, most benchmarks used in academic research do not explicitly try to capture such real-world input noises~\cite{Naplava2021UnderstandingMR}, leaving the doubt whether models performing well on standard clean test sets can transfer well onto real-world user-generated data.

To evaluate the performance on noisy data for languages like English, existing work typically generates typos via character-level perturbation such as randomly sampled or adversarial character swap or deletion ~\cite{Belinkov2018SyntheticAN,Pruthi2019CombatingAM,Jones2020RobustEA,Ma2020CharBERTCP}, automatic back-translation and speech conversion~\cite{Peskov2019MitigatingNI,Ravichander2021NoiseQACS}. 
However, there are many factors not considered in the automatic approaches, for example, the keyboard design of users' devices and speakers' phonetic and phonological variations. These overlooked factors have a large impact on the types of noises possible in keyboard and speech inputs. 
One notable exception to the above is NoiseQA~\cite{Ravichander2021NoiseQACS}. Apart from automatic approaches, they also collected test sets with noises produced by annotators. Their dataset only considered the question answering task and is only in English.

\begin{table}[t]
\small
\begin{center}
\setlength{\tabcolsep}{2mm}{
\begin{tabular}{ c | c }
\toprule
 \multirow{3}{*}{Original}  &   \textcolor{blue}{花}呗怎么不能\textcolor{orange}{提}\textcolor{orange}{额}了 (1a) \\
&   \textcolor{blue}{hu\={a}} bei z\v{e}n me b\`{u} n\'{e}ng ~\textcolor{orange}{t\'{i} \'{e}} le \\
 & \textit{Why can't I raise my quota on HuaBei?} \\
\midrule
 \multirow{2}{*}{Keyboard}  & 
花呗怎么不能\textcolor{orange}{贴}了 (1b) \\
 & hu\={a} bei z\v{e}n me b\`{u} n\'{e}ng ~\textcolor{orange}{ti\={e}} le \\
 \midrule
 \multirow{2}{*}{Speech}  & 
\textcolor{blue}{画}呗怎么不能\textcolor{orange}{提饿}了 (1c) \\
& \textcolor{blue}{hu\`{a}} bei z\v{e}n me b\`{u} n\'{e}ng ~\textcolor{orange}{t\'{i} \`{e}} le  \\
 \bottomrule
\end{tabular}}
 \caption{An example of our crowd-sourced keyboard and speech noises. The original question comes from AFQMC~\cite{CLUE}. We also present the Pinyin transliteration of the text. Colors indicate the original and corresponding mis-entered characters.}
 \label{tab:example}
\end{center}
\end{table}


In this paper, we focus on Chinese instead and present a multi-task benchmark with \textbf{\textsc{RE}}alistic \textbf{\textsc{A}}nd \textbf{\textsc{D}}iverse \textbf{\textsc{I}}nput \textbf{\textsc{N}}oise, named \textbf{\textsc{Readin}}.
Compared to the case of English, Chinese input noises have very different patterns due to the very different nature of the two languages. 
Chinese is a pictographic language without morphological inflections that are common in Indo-European languages. Also, the tone system is a unique and integral part of Chinese phonology but not in English. 
Such differences cause different types of input noises in both keyboard typing and speech input.
 To comprehensively study the effect of real-world noises, we cover four diverse tasks: paraphrase identification, machine reading comprehension, semantic parsing (text2SQL) and machine translation, all of which represent important real-life applications. 

 We consider noises occurring in two widely used Chinese input methods, keyboard input and speech input, and provide an example in Table~\ref{tab:example}. 

For keyboard input, Chinese users need to use an input method editor (IME) to convert the raw transliteration\footnote{There are also IME that convert radical sequences into characters. We focus on transliteration-based IME in this paper (in particular the Pinyin input method) since it's more commonly used among Chinese users~\cite{Fong2012CHINESEIM}.} sequences into Chinese characters. In such cases, noises can either occur in the transliteration input, or occur when users are choosing the intended word from the candidate list suggested by the IME. It is different from the case of English where typos and spelling variations are expected to happen on the character level. 
The noise patterns are further coupled with the typing habits of individual users, for example, whether they type the full Pinyin transliteration or just the abbreviations results in different noise patterns. 
In order to capture these nuances, we recruit annotators with different typing habits and instruct them to use different IMEs for typing.

For speech input, noises could arise when the speakers' accents or background noises lead to failures of the post-processing automatic speech recognition (ASR) systems. 
To capture these, we recruit 10 speakers from different regions of China to cover diverse accents and use a commonly used Chinese commercial ASR system for post-processing. For instance, in Table~\ref{tab:example}, the speech noise occurs because the speaker has different tones in their accent, leading the ASR system to produce different characters than the original ones. 
Ensuring that models are robust across these accent variations has important implications for fairness.

We take many additional measures in the annotation process in order to capture the real-world input noise distribution, as detailed in Section~\ref{sec:annotation}. In Section~\ref{sec:dataset}, we provide more statistics and analysis of the collected data. In Section~\ref{sec:baselines}, we train strong baseline models on the clean training data and test the models on our \textsc{Readin} test sets. The results indicate that these models suffer significant performance drops on the real-world input noises, leaving ample room for future improvement.

\section{Annotation Process}
\label{sec:annotation}

Our annotation asks crowdworkers to re-enter clean test data from these existing NLP datasets. Our goal is to induce realistic and diverse input noises in the annotation. 
We collect data using two different types of input methods: keyboard (Pinyin) input and speech input, both are commonly used among Chinese users~\cite{Fong2012CHINESEIM}. 
All examples are annotated with both input methods and we keep two separate tracks for data collected with these two different input methods. 
In the following subsections, we first introduce the four tasks and the original datasets that our annotations are based on, and then introduce the annotation process for keyboard input and speech input respectively. 


\subsection{Tasks and Original Datasets}
\label{sec:tasks}


\paragraph{Paraphrase Identification} is a binary classification task that aims to determine whether the given sentence pair are paraphrases. We use the AFQMC dataset~\cite{CLUE} as the original source for annotation, where the data come from customer services in the financial domain. The original dataset is unbalanced (with more negative pairs than positive), we down-sample the negative examples to make the training and dev sets balanced, and we report the accuracy separately for positive pairs and negative pairs. During annotation, we annotate both sentences in each sentence pair since in reality both sentences could be user-generated.

\paragraph{Machine Reading Comprehension} gives the model passage-question pairs and asks the model to output the correct answer. We choose a span-extraction MRC dataset CMRC2018~\cite{Cui2019CMRC2018} as the original data source. We use answer string exact match as the evaluation metric. During annotation, we only annotate the questions and keep the passages clean. This simulates the realistic setting where users enter their queries potentially with typos.

\paragraph{Semantic Parsing} requires the model to convert natural language queries into logical forms. We use the CSpider dataset~\cite{CSpider} which is a dataset for the natural language to SQL query task and is the Chinese version of the Spider dataset~\cite{Spider}. We use exact match as the metric. During annotation, we annotate the natural language questions to induce typos and use the original SQL queries as the gold reference. 

\paragraph{Machine Translation} requires the model to translate the input in the source language into the target language. We use the news translation shared task from WMT2021~\cite{WMT21} as our original data source. Following the standard practice of the MT community, we use SacreBLEU~\cite{Post2018ACF} to compute the BLEU score as the metric. During annotation, we only annotate the Chinese sentence and preserve the original English translation as the gold reference.

\begin{figure}[t]
\centering
\includegraphics[width=0.45\textwidth]{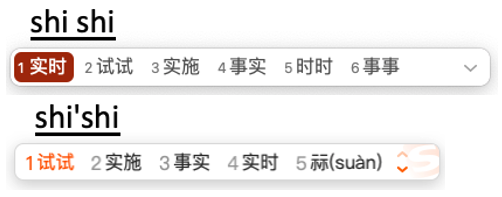}
\caption{A screenshot of two different Pinyin IMEs. Given the exact same Pinyin input (\textit{``shi shi''}), different IMEs suggest different words in different orders for users to select from. We use three different IMEs in keyboard annotation for wider coverage.}
\label{fig:ime}
\end{figure}

\subsection{Pinyin Input Annotation}

We present each annotator with a set of input data and ask them to re-type with the Pinyin input method. We implement the following restrictions in the annotation.\footnote{We also record the typing interface during the annotations to facilitate future analysis.}

\paragraph{Different IMEs} There are many commercial IME softwares available for the Pinyin input method. To maximize diversity, every input sentence is annotated by three different annotators, where each annotator uses a different IME software. We specified three commonly-used commercial Pinyin IMEs: Microsoft\footnote{\url{https://en.wikipedia.org/wiki/Microsoft_Pinyin_IME}}, QQ\footnote{\url{http://qq.pinyin.cn/}}, and Sogou\footnote{\url{https://pinyin.sogou.com/mac/}}. The main difference among these different IMEs is that when users type the same Pinyin transliteration input, 
different IME softwares suggest different candidate words and in different orders, as illustrated in Figure~\ref{fig:ime}. 
The use of different IMEs captures a wider range of possible typing noises. 

\paragraph{Speed Limit} Through our pilot run, we find that some annotators like to double-check their typed sequence.   
This is against our intention to collect more diverse noises for stress testing models, and we prefer to simulate cases where users may type in a much faster pace. 
Therefore, we set a speed limit of 40 characters per minute, which is the average rate of several runs of pilot annotation. 
We include a timer in the annotation pipeline and annotations with significantly slower typing speed are requested for re-annotation with a faster pace. 

\paragraph{Disallow Post-Editing} 
In pilot runs, we also find that some annotators like to correct their typos when they double-check their inputs, which again goes against our purpose. To complement the speed limit restriction, we also implement an additional constraint where post correction is not allowed in the annotation pipeline.


\subsection{Speech Input Annotation}

For speech input, we present each annotator with a set of input data and ask them to read and record them. The recordings are then converted to text data with ASR. We implement the following measures to ensure the diversity of speech input noises. 

\paragraph{Setup} To represent realistic settings, all recordings are done with mobile devices (the annotators' phones), with 16kHz sampling rate, which is high enough for ASR. We also instruct the annotators to record in environments with natural background noises, for example in their offices with some light background talking or street noises.

\paragraph{Diversity} There are large phonetic and phonological variations among different users especially since there are many accents across Chinese speakers. To capture such variation, we recruited a total of 10 different annotators for this speech input task (4 males and 6 females). They are selected from a larger pool of annotators through our trial run to maximally diversify accents. They come from different parts of China with different dialectic groups (more annotator details are in the appendix).
Their ages range from 32 to 64. We instruct the annotators to speak Mandarin while preserving their accents. Each input sentence is annotated by 3 different annotators from different dialectic groups to maximize diversity. 

\paragraph{ASR} The collected speech data are converted to text with a commercial automatic speech recognition (ASR) software iFlytek\footnote{\url{https://global.xfyun.cn/products/real-time-asr}}. We choose this commercial software because it is optimized for Mandarin and outperforms other open-source toolkits that we explored in the pilot run in terms of character-level error rates. We also release the raw audio recordings so that future work can explore using other alternative ASR choices as well.



Throughout the paper, we report results separately for the keyboard and speech noisy test sets for more fine-grained comparisons. 
We introduce more details of the annotated test sets in the next section. 

\begin{table}
\small
\begin{center}
\setlength{\tabcolsep}{2mm}{
\begin{tabular}{ l r r r }
\toprule
 Dataset & Train & Dev & Test  \\ 
  \midrule
  AFQMC &  18,000 & 2,000 & 4,317 \\
  CMRC2018 & 8,871 & 1,271 & 3,219 \\
  CSpider    & 7,500  & 1,159  & 1,034 \\
 WMT2021  &   --   &    --  & 1,948 \\  
 \bottomrule
\end{tabular}}
 \caption{Sizes of our four datasets.  For CMRC2018, we report the number of questions (multiple questions can correspond to the same passage). For WMT2021, we directly use the mBART50 model trained for multilingual translation without any additional finetuning on English-Chinese data, so there are no additional train or dev data involved.
 }
 \label{tab:stats}
\end{center}
\end{table}

\section{Dataset Overview}
\label{sec:dataset}

In this section, we analyse the annotated noisy test sets, including data statistics, our proposed metrics for robustness evaluation, a manual quality assessment of the annotated data as well as a qualitative analysis of the diverse types of input noises.


\subsection{Corpus Statistics}

The keyboard and speech noise data have the same sizes.\footnote{We performed some minimal filtering on the speech noise data to remove nonsensical outputs from ASR, which only involves about 50 examples in total and is omitted in the table.}
We only perform noise annotation on the test data and the training and dev sets remain clean. This serves our purpose to stress test models' robustness. Since the original datasets did not publicly release their test sets, we use their original dev splits as our test sets and we re-split the existing training data into our new train and dev splits, and we only annotate the test splits. We present the statistics of our data splits in Table~\ref{tab:stats}. 

To gauge the amount of noises in our annotated test sets, we report the character-level error rates for each noisy test set. Since the noise data could involve various changes like character deletion, insertion, or substitution, we use Levenshtein distance to measure the level of noise. Specifically, given a clean sentence $s$ and its annotated noisy version $t$, we define its error rate as: 

$$\mathrm{error} =  \frac{\mathrm{levenshtein}(s,t) }{\mathrm{len}(s) }$$

We measure the micro-average (average overall all annotations) as well as the worst-average (only consider the highest error rate annotation for each example) error rate across all three annotations over all examples. These two measures are further explained in the next section. 
The error rates are presented in Table~\ref{tab:error_rates}. We find that speech noises generally incur larger error rates except on CSpider, and in all cases, the error rates are well below 50\%.

\begin{table}[t]
\small
\begin{center}
\setlength{\tabcolsep}{2mm}{
\begin{tabular}{ c c c c c }
\toprule
    & \multicolumn{2}{c}{Keyboard} & \multicolumn{2}{c}{Speech} \\
     & Average & Worst & Average & Worst \\ 
 \midrule
    AFQMC       & 18.8 & 27.5 & 30.9 & 44.1 \\
    CMRC2018    & 17.4 & 26.9 & 25.1 & 38.1 \\
    CSpider     & 17.4 & 25.7 & 13.3 & 21.8 \\
    WMT2021     & 17.7 & 25.1 & 21.6 & 30.8 \\
 \bottomrule
\end{tabular}}
 \caption{Micro-average and worse-average error rates on our annotated test sets. Micro-average (`Average') is the mean of the average error rate among all three annotations for all examples. Worst-average (`Worst') takes the mean of the maximum error rate among all three annotations for all examples. }
 \label{tab:error_rates}
\end{center}
\end{table}

\subsection{Evaluation Metrics}

Apart from the individual metrics as introduced in section~\ref{sec:tasks}, we introduce two other benchmark-level metrics to account for the variations across the three different annotations per test example. 

Suppose for the $i$-th example, the performance of the model (by its task-specific metric) on the three typo annotations are $p^i_1, p^i_2, p^i_3$ respectively. We define the following two measures:

\paragraph{Micro-Average} takes the average of all performance across the three annotations, and then averages across all examples,
$$MA = \frac{1}{N}\sum_{i=1}^{N} (\frac{1}{3} \sum_{j=1}^{3} p^i_{j} ) $$
$$ = \frac{1}{3} (\frac{1}{N} \sum_{i=1}^{N} p^i_{1} + \frac{1}{N} \sum_{i=1}^{N} p^i_{2} + \frac{1}{N} \sum_{i=1}^{N} p^i_{3}). $$
In other words, this is equivalent to taking the average of the per-annotator performance.

\paragraph{Worst-Average} takes the minimum of the performance among all three annotations per average, and then averages across all examples, 
$$WA = \frac{1}{N}\sum_{i=1}^{N} \mathrm{min} (p^i_{1}, p^i_{2}, p^i_{3}).$$
This is a more challenging setting where we examine the worst-case performance across the annotation variations for each example.


\begin{table*}
\small
\begin{center}
\setlength{\tabcolsep}{2mm}{
\begin{tabular}{ c | c | l }
\toprule
 CMRC2018 &  Original  &  底特\textcolor{blue}{律}\textcolor{orange}{第二浸信会}教堂在哪里 (2a)   \\
 & & \textit{Where is Detroit's Second Baptist Church?} \\
 & Keyboard  &  底特律\textcolor{orange}{地热进行会}教堂在哪里 (2b) \\
& Speech  & 底特律\textcolor{orange}{第二情形会}教堂在哪里 (2c) \\
& Auto & 底特\textcolor{blue}{绿}第二浸信会教堂在哪里 (2d) \\
\midrule
CSpider  &  Original  &  
 \textcolor{blue}{8缸或}1980年前\textcolor{purple}{生}产的汽车的最大\textcolor{orange}{里程}是多少 (3a) \\
 & & \textit{What is the maximum mileage for an 8 cylinder or pre-1980 car?} \\
 & Keyboard  & 8缸或1980年前生产的汽车的最大\textcolor{orange}{历程}是多少 (3b) \\
& Speech & \textcolor{blue}{八港货}1980年前生产的汽车的最大里程是多少 (3c) \\
& Auto  & 
8缸或1980年前\textcolor{purple}{升}产的汽车的最大里程是多少 (3d) \\
\midrule
 WMT2021 &  Original  &  \textcolor{blue}{要尽力}\textcolor{purple}{防止}病\textcolor{olive}{毒}在社区进一步\textcolor{orange}{扩散} (4a) \\
 & & \textit{Try our best to fight against further spread of the coronavirus in the community.} \\
 & Keyboard  & 
 \textcolor{blue}{药剂量}\textcolor{purple}{发展}病毒在社区进一步\textcolor{orange}{开始} (4b) \\
& Speech  & \textcolor{blue}{要经历}防止病毒在社区进一步扩散 (4c) \\
& Auto  & 要尽力防\textcolor{purple}{指}病\textcolor{olive}{独}在社区进一步扩散 (4d) \\
 \bottomrule
\end{tabular}}
 \caption{More examples of different types of noises in \textsc{Readin}, in comparison with automatically constructed typos. The three examples are from three different datasets.}
 \label{tab:categorization}
\end{center}
\end{table*}

\subsection{Data Quality Analysis}

In order to analyze the quality of our annotated data, we design a human evaluation experiment. 
We compare our noisy test sets with the automatically constructed input noise test sets as in \citet{Si2021SubCharacterTF}. Specifically, they replace characters in the original sentences with randomly sampled homophones based on an existing Chinese homophone dictionary~\cite{OpenAttack}. We replicate their approach as a baseline and add an additional constraint that we only allow simplified Chinese characters in the character substitution process since our data focus on simplified Chinese. 

We aim to compare whether our crowdsourced noise data are more likely to occur in the real world. Towards this goal, we conduct a human preference selection experiment, where we present pairs of sentences to two annotators (different from the ones who did the noisy input annotation). Each pair consists of a sentence with automatic typos and another with our crowdsourced input noise, and the ordering is randomly shuffled for all pairs. We instruct the annotators to select the sentence that is more likely to occur in real user input settings (\textit{i.e.,} more plausible). 
We perform such annotation on 160 randomly sampled sentence pairs, for both keyboard input noises and speech input noises. 

We show some qualitative examples to compare our real-world noises and automatically constructed ones in Table~\ref{tab:categorization}, where we see that automatic noises involve substitutions that are unlikely to happen in real-world (for example only changing a single character ``毒" to ``独" in the word ``病毒" rather than mis-typing the entire word like human annotators tend to do). Quantitatively, we find that our crowdsourced keyboard input noises are preferred 87.5\% of the time as compared to automatic typos, and our speech input noises are preferred 86.3\% of the time compared to automatic typos (the results are averaged over two annotators). These results suggest that our crowdsourced noisy data are much more plausible than automatic typos.

\subsection{Diversity Analysis}

To understand the diversity of the noise patterns in our annotated data, we first present some qualitative case studies.
We present sampled examples in Table~\ref{tab:categorization} showing a wide range of noise patterns. We traced back to the annotation recordings to better understand how these noises arise during typing. 
In example (3b), ``里程'' and ``历程'' have the same Pinyin transliteration and the annotator chose the wrong word on the IME ; in example (4b), the annotator typed the abbreviation \textit{``y j l''} for \textit{``yao jin li'' (``要尽力'')}, which turned into \textit{``yao ji liang'' (``药剂量'')} due to wrong word selection (these two words have the same abbreviation);
in example (2b), 
the annotator mis-typed the Pinyin input by swapping \textit{``er'' (``二'')} to \textit{``re'' (``热'')}.

For speech input data, we listened to some sampled raw recordings and found that different annotators have vastly different accents leading to various noise patterns. 
The speech noise (1c) in Table~\ref{tab:example} shows an example where the first tone (`花' [hu\={a}]) is pronounced as the fourth tone (`画' [hu\`{a}]); in example (2c), \textit{``jin xin'' (``浸信'')} is pronounced as \textit{``qing xing'' (``情形'')}. The noises arise when these accent variations lead to corresponding characters through ASR post-processing.
Additionally, we found that the text data produced by the ASR system sometimes have a language modeling effect where the original words are replaced with more likely substitutes for better coherence (similar to the finding in~\citet{Peskov2019MitigatingNI} on English ASR). 
For example, in example (3c), \textit{``8缸或'' (``b\={a} g\={a}ng hu\`{o}'')} is converted to \textit{``八港货'' (``b\={a} g\v{a}ng hu\`{o}'')}. 

Quantitatively, we performed an additional annotation on 240 sampled keyboard input examples from six different annotators. We find that \textsc{Readin} examples cover different typing habits and noise patterns. For example, 69\% of the time annotators type the full Pinyin sequences while in 31\% cases annotators only type the abbreviated sequences; 56\% of these noises are due to selection errors (where the Pinyin input is right but the annotators selected the wrong word from IMEs) while the other 44\% are due to wrong Pinyin input.~\footnote{More details are in the Appendix.}

Overall, our analysis highlights that \textsc{Readin} covers realistic and diverse input noises, posing greater challenges for existing models.

\begin{table*}[t]
\small
\begin{center}
\setlength{\tabcolsep}{3mm}{
\begin{tabular}{ c c c c c c c c c c }
\toprule
    & \multicolumn{3}{c}{AFQMC (pos)} & \multicolumn{3}{c}{AFQMC (neg)}  & \multicolumn{3}{c}{CMRC2018} \\
     & Clean & Average & Worst & Clean & Average & Worst  & Clean & Average & Worst \\ 
 \toprule
 \multicolumn{10}{c}{\textit{Keyboard}} \\
 \midrule
    RoBERTa-wwm & 78.92 & 42.75 & 15.17 & 65.75 & 81.87 & 65.85 & 69.78 & 60.84 & 46.69  \\
    w/ ADA & 76.76 & 48.31 & 19.88 & 63.50 & 76.56 & 58.23 & 59.30 & 53.04 & 42.00 \\
    w/ Word Correction & 78.92 & 39.96 & 12.78 & 65.75 & 82.91 & 67.29 & 69.78 & 60.84 & 46.69 \\
    \midrule
    MacBERT & 80.04 & 48.33 & 18.83 & 62.09 & 76.77 & 58.29 & 67.69 & 56.71 & 41.29  \\
     w/ ADA & 77.88 & 53.21 & 24.66 & 64.30 & 74.41 & 55.34 & 59.24 & 54.05 & 43.99 \\
    w/ Word Correction & 80.04 & 44.52 & 16.22 & 62.09 & 78.51 & 60.41 & 67.69 & 56.72 & 41.29 \\
\midrule 
 \multicolumn{10}{c}{\textit{Speech}} \\
  \midrule
    RoBERTa-wwm & 78.92 & 27.75 & 5.68 & 65.75 & 87.80 & 73.81 & 69.78 & 55.97 & 40.73  \\
     w/ ADA & 76.76 & 39.76 & 13.30 & 63.50 & 78.26 & 58.93 & 59.30 & 48.32 & 36.35 \\
    w/ Word Correction & 78.92 & 27.75 & 5.68 & 65.75 & 87.80 & 73.81  & 69.78 & 55.97 & 40.73 \\
    \midrule
    MacBERT & 80.04 & 26.68 & 5.16 & 62.09 & 87.88 & 73.77 & 67.69 & 51.81 & 35.94 \\
     w/ ADA & 77.88 & 45.44 & 16.59 & 64.30 & 75.68 & 54.53  & 59.24 & 48.96 & 36.63 \\
    w/ Word Correction & 80.04 & 26.68 & 5.16 & 62.09 & 87.77 & 73.77 & 67.69 & 51.81 & 35.94 \\
 \bottomrule
\end{tabular}}
 \caption{Baseline performance on AFQMC and CMRC2018 test sets. We compare model performance on the original clean test set  (`Clean') and our new typo test sets. For results on typo test sets, we report both micro-average (`Average') and worst-average (`Worst') performance. For AFQMC, we report accuracy on positive and negative pairs separately. For CMRC2018, we report answer exact match.}
 \label{tab:nlu_results}
\end{center}
\end{table*}

\begin{table*}[t]
\small
\begin{center}
\setlength{\tabcolsep}{2mm}{
\begin{tabular}{c|ccccc|ccccc}
\toprule
    & \multicolumn{5}{c|}{CSpider} & \multicolumn{5}{c}{WMT2021}  \\ 
    \midrule
   &  & \multicolumn{2}{c}{Keyboard} & \multicolumn{2}{c|}{Speech} &  & \multicolumn{2}{c}{Keyboard} & \multicolumn{2}{c}{Speech} \\
   &  Clean & Average & Worst & Average & Worst & Clean & Average & Worst & Average & Worst \\ 
 \midrule
 DG-SQL / mBART50 & 44.87 & 28.85 & 11.99 & 33.40 & 24.18 & 23.19 & 16.35 & 9.37 & 16.74 & 10.82\\
w/ Word Correction & 44.87 & 30.24 & 13.73 & 33.40 & 24.47 & 23.19 & 17.59 & 10.24 & 16.89 & 10.97 \\
 \bottomrule
\end{tabular}}
 \caption{DG-SQL performance on CSpider and mBART50 performance WMT2021 test sets. We compare model performance on the original clean test set  (`Clean') and our new noisy test sets. For results on noisy test sets, we report both micro-average (`Average') and worst-average (`Worst') performance. For CSpider, we report exact match with the gold reference; for WMT2021, we report BLEU.}
 \label{tab:nlg_results}
\end{center}
\end{table*}

\section{Experiments}
\label{sec:baselines}

We benchmark several pretrained language models and examine whether their performance stays strong on \textsc{Readin}. 

\subsection{Baseline Setups}


We use RoBERTa-wwm ~\cite{Cui2021WholeWordMasking} and MacBERT~\cite{Cui2020MacBERT} as baselines for classification tasks. RoBERTa-wwm is a Chinese version of  RoBERTa~\cite{liu2018RoBERTa}, where whole-word-masking is used during pretraining. MacBERT is a modification to BERT~\cite{devlin2019BERT} where replaced word correction is used as a pretraining objective. Both of these models, like the original Chinese BERT, directly use the WordPiece~\cite{Wu2016WordPiece} tokenizer on Chinese characters. We use the base scale checkpoint for both models.

For machine translation, we adopt  mBART50~\cite{Tang2020mBART-50} as the baseline, which is a multilingual Transformer model that consists of 12 encoder layers and 12 decoder layers and is trained based on mBART~\cite{mBART} for multilingual translation. 
For semantic parsing, we use DG-SQL~\cite{Wang2021MetaLearningFD}, a competitive baseline on CSpider based on multilingual BERT~\cite{devlin2019BERT}. 

For experiments on AFQMC, CMRC2018, and CSpider, we finetune the pretrained checkpoints on the corresponding clean training sets. For WMT2021, we directly take mBART50 for inference without additional finetuning on Chinese-English parallel data since mBART50 itself is already trained on parallel translation data including Chinese-to-English. 


\subsection{Robustness Methods}

Apart from standard finetuning, we also experiment several robust training and data processing methods in order to assess how much can existing robustness methods solve our benchmark. We briefly introduce these methods below. 

\paragraph{Adversarial Data Augmentation}  ADA~\cite{Si2020BetterRB} is commonly used to enhance robustness against adversarial examples. We perform ADA by creating synthetic noisy training examples through random homophone substitution as in~\cite{Si2021SubCharacterTF} and add these examples to the original training examples. We double the number of total training examples through ADA. 

\paragraph{Typo Correction} Inspired by previous work that used a word recognition model to restore misspelled words in English~\cite{Pruthi2019CombatingAM}, we use a highly optimized commercial Chinese typo correction software\footnote{\url{https://console.xfyun.cn/services/text_check}} to pre-process data in READIN and then perform evaluation on the corrected data. We only perform this step on the noisy test sets, not the clean sets. 

\paragraph{SubChar Tokenization Models} \cite{Si2021SubCharacterTF} released a series of BERT-style models trained with SubChar tokenization, which use sub-character units such as radicals and syllables to compose Chinese characters. In particular, their SubChar-Pinyin model has the advantage of being robust to homophone typos. We adopt their model and also consider performing ADA on top of the SubChar-Pinyin model.

\begin{table}[t]
\small
\begin{center}
\setlength{\tabcolsep}{1mm}{
\begin{tabular}{c|ccccc}
\toprule
   &  & \multicolumn{2}{c}{Keyboard} & \multicolumn{2}{c}{Speech} \\
   &  Clean & Average & Worst & Average & Worst  \\ 
 \midrule
Subword & 75.81 & 49.63 & 22.03 & 42.21 & 19.31 \\
w/ ADA & 69.76 & 49.39 & 25.67 & 46.35 & 22.97 \\
\midrule
SubChar-Pinyin & 73.99 & 50.88 & 23.42 & 45.24 & 21.21  \\
w/ ADA &  73.73 & 54.16 & 29.43 & 52.93 & 28.06 \\
 \bottomrule
\end{tabular}}
 \caption{Finetuning results of BERT models trained with subword and SubChar tokenizers on the AFQMC (pos) subset. SubChar models are more robust than subword models, especially after performing  data augmentation.}
 \label{tab:subchar_results}
\end{center}
\end{table}

\subsection{Results}

We present results of the baseline models in Table~\ref{tab:nlu_results} (for NLU tasks) and Table~\ref{tab:nlg_results} (for NLG tasks). We highlight several main findings below.

\paragraph{Input Noises Cause Large Drops} 
We first compare performance of the same models on the clean test sets and the noisy test sets. 
We see a clear trend that model performance drops significantly when evaluated on the noisy test sets as compared to the clean test sets. As expected, the worst-average performance is much worse than the micro-average, showing that robustness across annotator variations is challenging. Moreover, we find that speech noises cause larger performance drops than keyboard noises (except on CSpider), which corresponds to the character error rates of these different test sets (Table~\ref{tab:error_rates}). 

One notable result is on AFQMC, where we observe drastic performance drop on the positive paraphrase pairs but marginal drop or even performance increase for negative pairs. The reason is that models are exploiting spurious correlation in the training data such as lexical overlap as cues for positive pairs~\cite{HANS,PAWS}. When we introduce input noises to the data, the lexical overlap decreases, thus models exploiting spurious features become more likely to predict negative labels. 
Better performance on the positive examples in AFQMC (without significant sacrifice on the clean tests) can be taken as a sign for better robustness.
We also present results on AFQMC as measured by the F1 metric in the appendix, and the results also indicate a drop in F1 on the noisy tests.

\paragraph{Robustness Methods Have Inconsistent Gains}
For the adversarial data augmentation (ADA) and word correction pre-processing methods, we find that they have inconsistent gains on different datasets. For example, ADA improves performance on the noisy test sets on the AFQMC (pos) set, but not on the CMRC2018 dataset. On the other hand, word correction improves performance on the keyboard noise test sets of CSpider and WMT2021, but not on the other datasets. 

\paragraph{SubChar Tokenization Helps} 
Lastly, in Table~\ref{tab:subchar_results}, we show results for finetuning models with SubChar tokenization. We find that the SubChar-Pinyin model outperforms the Subword model (which uses conventional subword tokenization). Moreover, the gain is much larger after training SubChar-Pinyin with ADA.


\section{Related Work}

\paragraph{Spelling Errors} Previous works have recognized the impact of spelling and grammatical errors in multiple languages. Several typo and grammatical corpora have been collected~\cite{GithubCorpus}, notably by tracking Wikipedia edits~\cite{WikEd,JapWiki}. The major difference with our work, apart from the language used, is that we focus on real-world downstream applications with diverse input settings. There is also effort on spelling error correction (SEC)~\cite{Wu2013ChineseSC,SpellingCheck}. While SEC aims to restore the spelling errors, our goal is to make sure models perform well on downstream applications even in the existence of input noises. Applying an SEC model as pre-processing could be one way to improve performance on our \textsc{Readin} benchmark. Other alternatives for training robust models against spelling errors include noise-aware training~\cite{NAT} and learning typo-resistant representation~\cite{MOE,BERTRAM,Ma2020CharBERTCP}. We believe such modeling explorations to future work.

\paragraph{Linguistic Variations} Our \textsc{Readin} not only relates to spelling errors or typos, but also related to linguistics variations especially in terms of phonological variations. Previous works have examined linguistic variations such as non-standard English~\cite{Tan2020ItsMT,Tan2020MindYI,Groenwold2020DatsWI} and dialect disparity~\cite{Ziems2022VALUEUD}. Such works have important implications for building equatable NLP applications especially for minority language groups in the society. Yet, such effort is absent in Chinese NLP and our benchmark is a first attempt towards incorporating linguistic variations in model evaluation. 

\paragraph{Adversarial Robustness} Works in the adversarial robustness often involved adversarially optimized character or word perturbations in an attempt to minimize model performance~\cite{Ebrahimi2018OnAE,Ebrahimi2018HotFlipWA,Jones2020RobustEA}. Corresponding defenses have also been proposed such as adversarial training or data augmentation~\cite{Belinkov2018SyntheticAN,Si2020BetterRB,Si2021BenchmarkingRO}. Our work differs from this adversarial robustness line of work because we are not measuring worst-case attacks, but rather more realistic input noises that would actually occur in real-world user-generated inputs. 



\section{Conclusion}

In this work, we present \textsc{Readin} - the first Chinese multi-task benchmark with realistic and diverse input noises. Our annotation is carefully designed to elicit realistic and diverse input noises for both keyboard Pinyin input and speech input. Through both quantitative and qualitative human evaluation, we show that our crowdsourced input noises are much more plausible and diverse than existing automatically created ones. Our experiments on strong pretrained language model baselines show that models suffer significant drops on our noisy test sets, indicating the need for more robust methods against input noises that would happen in the real world.

\section*{Ethics and Broader Impact}

We use this additional section to discuss potential ethical considerations as well as broader impact of our work. 

\paragraph{Ethical Consideration}
This work involves human annotation. We made sure that all annotators are properly paid. We discussed extensively with all annotators involved to set a compensation that all agree on before starting the annotation, and the total cost of annotation for the project is about 30K RMB. We also explicitly informed all annotators about how the collected data will be used and made adjustments in the data collection and release protocol to avoid any privacy concerns. Overall, we believe that there is no harm involved in this project's annotation jobs.

\paragraph{Positive Societal Impact}
This project tackles the real-world problem of input noises. We believe that our work will have a positive societal impact because we collected test data from annotators with diverse backgrounds. Our benchmark will facilitate the development of models that can perform well across all these variations, which has important implications to ensure the accessibility of our language technologies to users from diverse backgrounds. This fairness and inclusion aspect is often under-valued in the Chinese NLP community and we hope that our work can push the community to put more work on this front.

\paragraph{Limitations}
While we tried our best to maximize the diversity and coverage of our benchmark, it is practically impossible to cover all possible input noises. We acknowledge aspects that we did not get to cover, for example, the impact of different input devices (phones, tablets, as compared to keyboards used in our annotation). Also, while we tried to re-construct the real-world input settings as much as possible, there may still be subtle differences between real-world input and our annotation process, for example, we posed speed limits during the keyboard input annotation and this may not capture exactly how users type in real applications. We encourage future work to consider how to increase the coverage of such benchmarks and also possible innovations in the data collection procedure to collect fully realistic user data.  

\section*{Acknowledgement}

We thank members of the UMD CLIP lab for their helpful feedback. 
This work is supported by the National Key R\&D Program of China (No. 2020AAA0106502) and Institute Guo Qiang at Tsinghua University.

\paragraph{Author Contributions}
Chenglei and Zhengyan conducted the data collection; Yingfa ran most of the computational experiments. Chenglei, Zhengyan, and Yingfa did data cleaning and post-processing. Chenglei and Zhengyan wrote the initial draft. Xiaozhi provided advice for the data collection. Yingfa, Xiaozhi, and Zhiyuan Liu edited and improved the writing of the paper. Zhiyuan Liu and Maosong Sun supervised the project and provided helpful high-level guidance.

\bibliography{anthology,custom}

\clearpage

\appendix

\section{Appendix}
\label{sec:appendix}


\subsection{Annotator Details}

We provide more details about the speakers for our speech input annotation in Table~\ref{tab:speakers}. The hometowns also represent their dialectal groups. Our selected annotators represent a wide range of dialectal groups in China.

\begin{table}[h]
    \small
    \begin{center}
        \setlength{\tabcolsep}{3mm}{
        \begin{tabular}{ c c c}
            \toprule
                Age & Gender & Hometown (Accent)  \\ 
             \toprule
             Male & 35 & Harbin, Heilongjiang \\
             Male & 64 & Loudi, Hunan \\
             Female & 43 & Hefei, Anhui \\
             Male & 45 & Zhangjiakou, Hebei \\
             Male & 32 & Datong, Shanxi \\
             Female & 43 & Loudi, Hunan \\
             Female & 57 & Changde, Hunan \\
             Female & 32 & Shijiazhuang, Hebei \\
             Female & 33 & Guangyuan, Sichuan \\
             Female & 36 & Zigong, Sichuan \\
             \bottomrule
        \end{tabular}}
         \caption{Details about the ten speakers that performed the speech input annotation.}
         \label{tab:speakers}
    \end{center}
\end{table}

\subsection{AFQMC F1 Results}

We present evaluation results on AFQMC with the F1 metric in Table~\ref{tab:f1}. We can see significant performance drops on the noisy test sets. We prefer to report accuracy numbers for the positive and negative examples separately in the main paper because they better capture the different performance patterns for the positive and negative examples. 

\begin{table}[h]
    \small
    \begin{center}
        \setlength{\tabcolsep}{3mm}{
        \begin{tabular}{ c c c }
            \toprule
                & Clean & Average \\ 
             \toprule
             \multicolumn{3}{c}{\textit{Keyboard}} \\
             \midrule
                RoBERTa-wwm & 68.04 & 59.96 \\
                MacBERT     & 69.20 & 60.63 \\
            \midrule 
             \multicolumn{3}{c}{\textit{Speech}} \\
              \midrule
                RoBERTa-wwm & 69.19 & 46.89 \\
                MacBERT     & 68.04 & 43.90 \\
             \bottomrule
        \end{tabular}}
         \caption{Macro-F1 performance of baseline models on the entire AFQMC test set.}
         \label{tab:f1}
    \end{center}
\end{table}

\subsection{Noise Type Annotation}

\begin{table}[t]
\small
\begin{center}
\setlength{\tabcolsep}{2mm}{
\begin{tabular}{ c c c }
\toprule
     & Full & Abbr \\ 
 \midrule
    Wrong Input  & 29.8\% & 14.3\% \\
    Wrong Selection  & 39.3\% & 16.7\% \\
 \bottomrule
\end{tabular}}
 \caption{Noise breakdown of sampled Pinyin input examples. We categorise the noises into four types based on whether they are types as full Pinyin sequences (Full) or abbreviations (Abbr) and whether the noises are due to wrong input or word selection.}
 \label{tab:error_breakdown}
\end{center}
\end{table}

To better understand the different noise patterns and diversity of the keyboard noise data, we perform an additional human annotation on two keyboard input subsets in \textsc{Readin}: AFQMC and WMT2021. From each dataset we examine the annotation recording of 40 sentences from different annotators. Since there are three annotators for each dataset (each using a different IME), this results in a sample size of 240 sentences for this human annotation. The authors of this paper performed this annotation task by categorising the noises in these sampled inputs into four categories detailed below. 

We note that the annotators have two different typing habits: they either input the full Pinyin sequence or the abbreviations (\textit{e.g.}, just typing the first syllables of each character). Orthogonal to these different typing habits, the noises have two different sources: they either occur because the input Pinyin sequence is wrong or the input sequence is right but the original annotators selected the wrong word in the IME. The combination of these two typing habits and error sources results in the four noise types listed in Table~\ref{tab:error_breakdown}. We follow such a scheme for error breakdown because these categories represent very different noisy input patterns and may pose different challenges for the models. 

From Table~\ref{tab:error_breakdown}, we can see that wrong word selection is more common than wrong input sequences, and typing in full is more common than typing abbreviations. Moreover, there are a significant number of examples from each category, confirming the diversity of the noise patterns in the Pinyin input annotations.

\end{CJK*}
\end{document}